\documentclass{article}
\usepackage{ijcai22}

\usepackage{times}

\usepackage{soul}
\usepackage{url}
\usepackage[hidelinks]{hyperref}
\usepackage[utf8]{inputenc}
\usepackage[small]{caption}
\usepackage{graphicx}
\usepackage{amsmath}
\usepackage{booktabs}
\usepackage{amsthm}
\usepackage{amssymb}
\usepackage{algorithm}
\usepackage{algorithmic}
\usepackage{bm}
\usepackage[authoryear,longnamesfirst]{natbib}
\title{Graph-level Protein Representation Learning by Structure Knowledge Refinement}

\author{
    Ge Wang, Zelin Zang, Jiangbin Zheng, Jun Xia, Stan Z. Li\\
}

\begin{document}

\maketitle

\begin{abstract}
    This paper focuses on learning representation on the whole graph level in an unsupervised manner. Learning graph-level representation plays an important role in a variety of real-world issues such as molecule property prediction, protein structure feature extraction and social network analysis. The mainstream method is utilizing contrastive learning to facilitate graph feature extraction, known as Graph Contrastive Learning (GCL). GCL, although effective, suffers from some complications in contrastive learning, such as the effect of false negative pairs. Moreover, augmentation strategies in GCL are weakly adaptive to diverse graph datasets. Motivated by these problems, we propose a novel framework called Structure Knowledge Refinement (SKR) which uses data structure to determine the probability of whether a pair is positive or negative. Meanwhile, we propose an augmentation strategy that naturally preserves the semantic meaning of original data and is compatible with our SKR framework. Furthermore, we illustrate the effectiveness of our SKR framework through intuition and experiments. And the experimental results on the tasks of graph-level classification demonstrate that our SKR framework is superior to most state-of-the-art baselines.
\end{abstract}

\section{Introduction}

Graph is a powerful tool to represent diverse types of data including small molecules~\cite{huber2007graphs}, protein-protein interaction~\cite{bu2003topological} and social networks~\cite{newman2004finding}. Unlike Euclidean data such as image data, graph data whose underlying structure is a non-Euclidean space describes coupling relationships between individual units in a large framework through assigning properties to the nodes and the edges connecting them. Recently, With the development of graph convolutional network~\cite{kipf2016semi}, people pay increasingly attention to extending deep learning approaches for graph data. Due to the expensiveness of handcrafted annotation, self-supervised learning is emerging as a new paradigm for extracting informative knowledge from graph data. 


As a mainstream method in graph self-supervised learning, Graph Contrastive Learning (GCL) uses augmentation strategies to generate multi-views of instances and use the information of differences and sameness between those views to learn the intrinsic representations. Specifically, two views generated from the same instance are treated as a positive pair, while two views generated from different instances are treated as a negative pair. The main idea of constructing pretext task for GCL to learn representation is to maximize the agreement of positive pairs and minimize the agreement of negative pairs.~\cite{wu2021self} However, under such paradigm, there are two issues: 

\textbf{(1) Harmful effect of false negative pairs.} Without access to labels, treating views from different instances as negative pairs implicitly accepts that views in a negative pair may, in reality, actually have the same semantic (i.e.same label). We call such pair as false negative pair. Thus we can decompose negative pairs into two parts: true negative pairs and false negative pairs~\cite{robinson2020contrastive, chuang2020debiased}. Views in each true negative pair have different semantic, and views in each false negative pair has same semantic. By minimizing agreement of true negative pairs do encourage representation learning, however minimizing agreement of false negative pairs will enforce model to push representation with same semantic away and lead to convergence difficulties~\cite{huynh2022boosting}. 

\newcommand{\tabincell}[2]{\begin{tabular}{@{}#1@{}}#2\end{tabular}}

\begin{table}[tb]
    \footnotesize
    \centering
    \vspace{4pt}
    \scalebox{0.9}{
    \renewcommand{\arraystretch}{1.2}
    \begin{tabular}{|c|c|c|}
        \hline
        \tabincell{c}{Augmentation \\ Strategy}         & Example       & Prior Knowledge\\ \hline
        \tabincell{c}{Feature-based \\ augmentation}   &\tabincell{c}{Attribute masking \\ Attribute shuffling}     & \tabincell{c}{Attribute change does \\ not alter semantic.} \\ \hline
        \tabincell{c}{Structure-based \\ augmentation} &\tabincell{c}{Edge perturbation \\ Node Insertion}    &\tabincell{c}{Structure change does \\ not alter semantic} \\ \hline
        \tabincell{c}{Sampling-based \\ augmentation}   &\tabincell{c}{Random walk sampling \\ Uniform sampling}    & \tabincell{c}{Local structure can \\ hint the full semantic.}     \\
        \hline
    \end{tabular}}
    \caption{Overview of graph augmentation strategies.}
    \vspace{-8pt}
    \label{tab_augmentation}
\end{table}

\textbf{(2) Weak adaptability of graph augmentation strategy.} Due to the inherent non-Euclidean property of graph data, it is difficult to directly apply the image augmentation strategies to graph data. In different levels, augmentation strategies of graph can be approximately summarized into three categories: feature-based augmentation, structure-based augmentation and sampling-based augmentation (Tab.~\ref{tab_augmentation}). Because graph datasets are abstracted from diverse fields, there may not be universally appropriate data augmentation as those for image. Thus some graph augmentation strategies may be only suitable for certain datasets, and they may alter the semantic when augment some other datasets. In other words, the graph augmentation strategies rely on certain prior knowledge~\cite{you2020graph}, and are weakly adaptive to diverse graph datasets.

Motivated by these problems, we borrow the main idea of GCL and propose a novel method called Structure Knowledge Refinement (SKR). Instead of treating pair either as positive or negative, we use probability to describe the relationship of views in a pair. Thus the sign of a pair is a fuzzy variable, and we can use fuzzy cross-entropy~\cite{luukka2011feature} as objective of our SKR method. Under such setting, our SKR model can automatically appeal views with same semantic and repeal views with different semantic, thus achieves the purpose of refining data structure. Additionally, by analyzing conventional method of deriving graph-level representation from node-level representation, we propose a graph-level representation augmentation strategy with no need of prior knowledge, thus further improves refining data structure.

We summarize out contributions as follows:
\begin{itemize}
  \vspace{-2pt}
  \item We propose a framework for graph-level representation learning called Structure Knowledge Refinement (SKR), which can automatically appeal views with same semantic and repeal views with different semantic.
  \vspace{-2pt}
  \item We propose a generalizable graph-level representation called Dirichlet Pooling, which can naturally preserve the semantic and is strongly adaptive to various graph datasets.
  \vspace{-2pt}
  \item We illustrate the effectiveness of our method through experiment. The experimental results show that SKR outperforms current state-of-the-art graph-level representation learning method.
\vspace{-4pt}
\end{itemize}
\section{Related Work}


The field of graph representation learning has grown at an incredible pace over the past decades. It derives graph embedding through various methods, and we separate them into three parts:

\textbf{Kernel based Graph Embedding.} Graph kernel is a kernel function that computes an inner product on graphs. It can be intuitively understood as a function measuring the similarity of pairs of graphs. Current Popular graph kernels are graphlets~\cite{prvzulj2007biological,shervashidze2009efficient},  random walk and shortest path kernels~\cite{borgwardt2005shortest}, and the Weisfeiler-Lehman subtree kernel~\cite{shervashidze2011weisfeiler}. Furthermore, deep graph kernels~\cite{yanardag2015deep} and multiscale Laplacian graph kernels~\cite{kondor2016multiscale} have been proposed with the goal to redefine kernel functions to appropriately capture sub-structural similarity at different levels.

\textbf{Skip-gram Based Graph Embedding.} skip-gram model for graph are central many popular word-embedding methods~\cite{mnih2013learning,mikolov2013distributed}. Word2vec~\cite{mikolov2013efficient} is an unsupervised algorithm which obtains word representations by using the representations to predict context words (the words that surround it). Doc2vec~\cite{le2014distributed} is an extension of the continuous Skip-gram model that predicts representations of words from that of a document containing them.

\textbf{Contrastive Learning Based Graph Embedding.} Graph contrastive learning is a mainstream method in unsupervised graph representation. The pretext task of GCL is to maximize the agreement of positive pairs and minimize the agreement of negative pair. More specially, representations of pairs are treats as random variables and mutual information is used to describe the relationship of them. InfoGraph~\cite{sun2019infograph} extends deep InfoMax~\cite{hjelm2018learning} and contrasts graph and nodes in the graph to gain graph-level representations. GraphCL~\cite{you2020graph} applies a series of graph augmentations randomly selected from node dropping, edge perturbation, attribute masking and subgraph sampling to generate an augmented graph then contrasts original graph and augmented graph to gain graph-level representations. AD-GCL~\cite{suresh2021adversarial} optimizes adversarial graph augmentation strategies used in GCL to enables GNNs to avoid capturing redundant information during the training. In GCL, another importance part is the selection strategies for negative samples. Conventional strategies uniformly select negative samples, which causes harmful effect of false negative pairs. To solve this, HCL~\cite{robinson2020contrastive} and DCL~\cite{chuang2020debiased} use rejection sampling to estimate the distribution of hard negative samples. Our method solve these issue by using probability to judge whether a pair is positive or negative then using fuzzy cross entropy~\cite{luukka2011feature} as objective to automatically attract views in positive pairs and repel views in negative pairs to gain graph-level representation.


\section{Framework}

\begin{figure*}[t]
    \centering
    \includegraphics[width=5.5in]{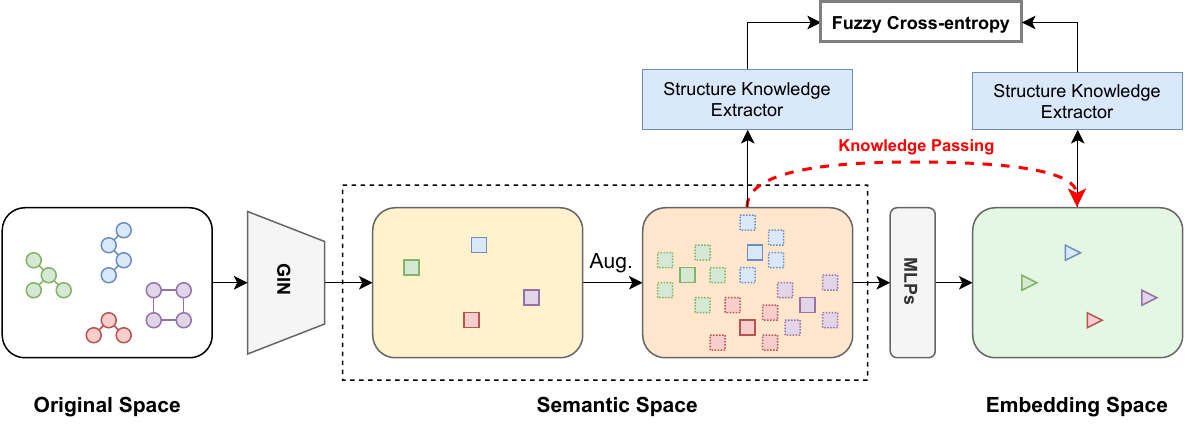}
    \vspace{-8pt}
    \caption{The framework of Structure Knowledge Refinement (SKR). Graph-level representations in semantic space are derived from graph data in original space by Graph Isomorphism Network (GIN), and augmented graph-level representations are generated by our semantic preserving augmentation strategy. Then semantic-space structure knowledge is obtained by structure knowledge extractor, and fuzzy cross-entropy is used to refine data structure in embedding space to derive better representations by passing semantic-space structure knowledge into embedding space.}
    \vspace{-8pt}
    \label{fig_framework}
\end{figure*}

To better illustrate our SKR framework, we use origin space, semantic space, and embedding space to present the space on which original graph data, semantic graph-level representation, and embedding graph-level representation lies respectively. The semantic graph-level feature is derived from original graph data passing through graph neural network, and embedding graph-level feature is obtained from semantic graph-level by using simple MLP structure (Fig.~\ref{fig_framework}). Then, we will from three aspects to introduce our framework:


\subsection{Architecture of SKR}

Our model architecture is similar to conventional GCL models (Fig.~\ref{fig_comparison}), so compared with those models, we don't need to learning extra training parameters. However, instead of basing on mutual infomation maximization principle, we use an intermediate semantic space to derive the probability of whether a pair is positive or negative, (i.e. if the distance of two samples are relatively close in semantic space, they are more likely to be a positive pair, and if the distance are quite far, they are more likely to be a negative pair), and those probabilities can well describe the structure of data in semantic space, thus we call them structure knowledge in semantic space. Because the distance metric needs to base on Euclidean space, we use graph isomorphism network (GIN)~\cite{xu2018powerful} to map non-euclidean updated node-level feature into semantic space. Then we use the structure knowledge from semantic space to help us obtain graph-level representation in embedding space by using cross-space loss. In this procedure, we also use augmentation in semantic space to further refine the data structure in embedding space.

\subsection{Augmentation Strategy of SKR}

In our framework, we use a novel augmentation strategy to enrich and refine structure knowledge in semantic space to enhance the feature extraction ability in embedding space. Unlike traditional augmentation strategy (e.g. node dropping, edge perturbation, attribute masking, sub-graph) requiring certain prior knowledge and generating data in original space (show in Table.~\ref{tab_augmentation}), our augmentation strategy generates augmented data in semantic space using the concept of combination mix-up and Dirichlet distribution . To further demonstrate our idea, let  $f^S_{G_i} \in R^{h_s}$ denote the representation of graph $G_i$ in semantic space; $f^S_{v_j} \in R^{h_S}$ denote the representation of node $v_j$ in graph $G_i$;  for graph $G_i$, it contains $|G_i|$ nodes $\{v_1, v_2, \cdots, v_{|G_i|}\}$. For graph-level representation learning, ones always use global add pooling or global mean pooling of node-level representation to express graph-level representation. In other words, for graph $G_i$, the graph-level representation can be written as

\begin{equation}
     f^S_{G_i}= \sum_{j=1}^{|G_i|}\frac{1}{|G_i|}f^S_{v_j}
\end{equation}

The main idea of our augmentation strategy is slightly perturbing the weight of each node-level representation $\frac{1}{|G_i|}$, in order to fulfill this idea, we can use Dirichlet distribution to randomly generate weights

\begin{equation}
    (\omega_1, \omega_2, \cdots, \omega_{|G_i|}) \sim \text{Dirichlet}(\underset{|G_i|}{\underbrace{\alpha, \alpha, \cdots, \alpha}})
\end{equation}
Thus, augmented graph-level representation can be written as

\begin{equation}
    \tilde{f}^S_{G_i}= \sum_{j=1}^{|G_i|}\omega_j f^S_{v_j}
\end{equation}

According to the property of Dirichlet distribution
\begin{equation}
    \begin{aligned}
        & \sum_{i=1}^{|G_i|}{\omega_i}=1 \ \ \ (\omega_i \geq 0)\\
        & \mathbb{E}[\omega_1]=\mathbb{E}[\omega_2]=\cdots =\mathbb{E}[\omega_{|G_i|}]=\frac{1}{|G_i|} \\
        & var(\omega_1)=\cdots=var(\omega_{|G_i|})=\frac{\alpha(\alpha|G_i|-\alpha)}{\alpha^2|G_i|^2(|G_i|\alpha+1)}
    \end{aligned}
    \label{property of Dirichlet}
\end{equation}
We can tuning the extent of perturbation by change hyper-parameter $\alpha$.

if $\alpha \rightarrow \infty$, $(w_1, w_2, \cdots, w_{|G_i|}) \rightarrow (\frac{1}{|G_i|}, \frac{1}{|G_i|}, \cdots,  \frac{1}{|G_i|})$, the augmented graph-level representation in semantic space is exactly same as naive graph-level representation in semantic space.

if $\alpha \rightarrow 0$, $(w_1, w_2, \cdots, w_{|G_i|})$ will approach to one-hot vector which means one entry's value is 1 meanwhile the other entry's value is 0, in this case,  the augmented graph-level representation in semantic space will degenerate into a node-level representation belong to corresponding graph in semantic space.

From analysis above, we can see by using this augmentation strategy, the augmented graph-level representation certainly lies in the region composed by corresponding graph's node-level representation, thus through use Dirichlet distribution our augmentation strategy can naturally preserve the semantic without any prior knowledge. Our augmentation strategy combines the properties of Dirichlet distribution and mean pooling, thus we name it Dirichlet pooling.

\subsection{Objective function of SKR}


Another key element in our framework is the cross-space loss which helps us to get the graph-level representations in embedding space through structure knowledge of augmented semantic graph-level representations. It can be separated into two steps: structure knowledge extractor and structure knowledge passing. 

For \textbf{structure knowledge extractor} part, graph-level representations in semantic space are entered into the structure knowledge extractor to calculate the pairwise distance matrix, then we derive the probability matrix by mapping distance into probability. To be more clear, let $\tilde{S}$ denote the pairwise distance matrix of all graph-level representations (including augmented and original) in semantic space $\tilde{S}_{ij}=||f^S_{G_i}-f^S_{G_j}||_2$, then the pairwise probability matrix $S$ in semantic space is calculated by mapping distance into $t^2$ distribution (a map from $[0, \infty)$ to $[0, 1]$) (Eq.~\ref{t^2 distribution}). 
\begin{equation}
        S = C_v(1+\frac{\tilde{S}}{\nu})^{-(\nu+1)}
\label{t^2 distribution}
\end{equation}
$C_v$ is the normalization factor of $t^2$ distribution, $\nu$ is the degree of freedom in $t^2$ distribution. In our framework we set $\nu$ to a quite large value, so it is equivalent to use normal distribution to map distance into probability. For the element $S_{ij}$ in probability matrix $S$, it means the probability that i-th sample and j-th sample have same semantic (i.e. the probability that they form a positive pair in semantic space). By mapping distance to $t^2$ distribution, we can describe the data structure more easily due to introducing non-linearity. And because the probability matrix $S$ can describe the data structure in semantic space, we call the probability matrix $S$ as structure knowledge in semantic space.

For the \textbf{structure knowledge passing} part, the fuzzy cross-entropy loss build the bridge between graph-level representation in semantic space and embedding space. And it is applied to make representation in embedding space include more useful information with the help of augmented structure knowledge in semantic space.  Our objective function based on fuzzy cross-entropy is defined as
\begin{equation}
    \mathcal{L}= -\sum_{i \not = j}[S_{ij}\log{E_{ij}}+(1-S_{ij})\log{(1-E_{ij})}]
\label{loss}
\end{equation}
Where $S_{ij}$ is the augmented structure knowledge in semantic space, $E_{ij}$ is the structure knowledge that needs to be refined in embedding space.


We will explain intuitively how this loss works, if graph-level representation i and graph-level representation j in semantic space are close to each other,  $S_{ij}$ will be close to 1 then first term will play a more importance role in loss, as result the corresponding representation in embedding space will attract each other and become closer and closer. In contrary, if two representations are far away from each other, the corresponding representation in embedding space will repel each other and become further and further. Consequently, the representations with same semantic in embedding space will become closer and closer, the representation with different semantic will going further and further. Thus, using fuzzy cross-entropy loss achieves the purpose of refining structure knowledge in embedding space through passing augmented structure knowledge in semantic space.

\subsection{Comparison of SKR and GCL}

\begin{figure}[t]
    \centering
    \includegraphics[width=3.3in]{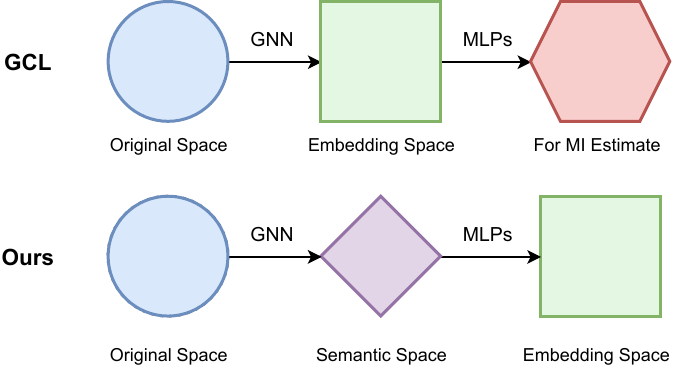}
    \vspace{-8pt}
    \caption{Comparison of SKR and GCL.}
    \vspace{-8pt}
    \label{fig_comparison}
\end{figure}

As shown in Fig.~\ref{fig_comparison}, our model are very similar to conventional GCL framework. Although SKR and GCL have same network architecture, they are based on different principle. 

For GCL, its output in embedding space is next to the original space, the role of MLPs is estimating mutual information, and the objective function of GCL is maximizing mutual information of positive pairs and minimizing mutual information of negative pairs.

For SKR, its output in embedding space is derived from an intermediate semantic space, the role of MLPs is refining data structure in embedding space to get better representations in embedding space, and the objective function of SKR is fuzzy cross-entropy which plays the role of passing augmented structure knowledge in semantic space into embedding space to refine the data structure in embedding space. And another key difference compared with GCL is that SKR use probability to judge whether a pair is positive or negative.

\section{Experiments}

In this section, we evaluate our SKR with a number of experiments. We first show the implementation of our SKR by pseudocode, then describe datasets and other settings (baseline and hyper-parameters). Next, we present the experimental results on graph classification. Last, we analyze our model via ablation study and sensitivity analysis.

\begin{algorithm}[hp]
\caption{Structure Knowledge refinement (SKR)}
\label{alg:algorithm}
\textbf{Input}: Graph data $\mathcal{G}=\{G_1, ..., G_{|\mathcal{G}|}\}$; GINs $g_{\theta}$, MLPs $h_{\phi}$, mean pooling $\mathcal{P(\cdot)}$, Dirichlet pooling $\mathcal{P_{D}(\cdot|\alpha)}$; training Epoch $T$, parameter of Dirichlet distribution $\alpha$, learning rate $\eta$ \\
\textbf{Output}: graph-level representations $f^E_{\mathcal{G}}=\{f^E_{G_1}, ..., f^E_{G_{|\mathcal{G}|}} \}$ \\
\begin{algorithmic}[1] 
\STATE Let $t=0$.\\
\STATE Initialize final representations $f^E_{\mathcal{G}}=h_{\phi}(\mathcal{P}(g_{\theta}(\mathcal{G})))$ \\

\WHILE{$t<T$; $t++$}
    \STATE Calculate structure knowledge of $f^E_{\mathcal{G}}$ $\rightarrow$ E (\ref{t^2 distribution})
    \STATE Calculate semantic representations $f^S_{\mathcal{G}}=\mathcal{P}(g_{\theta}(\mathcal{G}))$\\
    \STATE Do augmentation in semantic space $\tilde{f}^S_{\mathcal{G}}=\mathcal{P_{D}}(f^S_{\mathcal{G}}|\alpha)$
    \STATE Calculate structure knowledge of $f^S_{\mathcal{G}}$ and $\tilde{f}^S_{\mathcal{G}}$ $\rightarrow$ S (\ref{t^2 distribution})
    \STATE Calculate fuzzy cross entropy $\mathcal{L}$ (\ref{loss})
    \STATE Refine structure knowledge $E$ according to semantic structure knowledge $S$ to get better representations $f^E_{\mathcal{G}}$ \\ 
    $f^E_{\mathcal{G}}=h_{\phi-\eta \nabla_{\phi}L}(\mathcal{P}(g_{\theta-\eta \nabla_{\theta}L}(\mathcal{G})))$
    
\ENDWHILE
\STATE \textbf{return} solution
\end{algorithmic}
\end{algorithm}

\begin{table*}[ht]
    \footnotesize
    \centering
    \renewcommand{\arraystretch}{1.2}
    \begin{tabular}{c|ccccc|ccc|cc}
        \hline
        Domain &\multicolumn{5}{c|}{Social network} &\multicolumn{3}{c|}{Small molecules} &\multicolumn{2}{c}{Bioinformatics}  \\
        \hline
        Dataset &IMDB-B &IMDB-M &REDDIT-B &REDDIT-M5K &COLLAB &MUTAG &PTCMR &NCI1 &DD &PROTEINS \\ 
        \hline
        Graphs &1000 &1500 &2000 &4999 &5000 &188 &344 &4110 &1178 &1113 \\
        \hline
        Avg. Nodes &19.77 &13.00 &429.63 &508.52 &74.49 &17.93 &14.29 &29.87 &284.32 &39.06 \\
        \hline
        Avg. Edges &96.53 &65.94 &497.75 &594.87 &2457.78 &19.79 &14.69 &32.30 &715.66 &78.82 \\
        \hline
        Classes &2 &3 &2 &5 &3 &2 &2 &2 &2 &2 \\
        \hline
    \end{tabular}
    \caption{Summary of small molecules, bioinformatics and social networks from TU Benchmark Dataset~\cite{morris2020tudataset} used for unsupervised learning experiments. The evaluation metric for all these datasets is Accuracy.}
    \label{tab_datasets}
\end{table*}
        

\begin{table*}[tb]
    \footnotesize
    \centering
    \scalebox{0.9}{
    \renewcommand{\arraystretch}{1.2}
    \begin{tabular}{c|ccccc|ccc|cc}
        \hline
        Methods     &COLLAB       &IMDB-B       &IMDB-M       &REDDIT-B     &REDDOT-M5K      &MUTAG         &PTCMR        &NCI1         &DD           &PROTEINS  \\
        \hline
        SP           &-            &55.6$\pm$0.2 &38.0$\pm$0.3 &64.1$\pm$0.1 &39.6$\pm$0.2 &85.2$\pm$2.4  &58.2$\pm$2.4 &79.3$\pm$0.4 &74.5$\pm$0.2 &\textbf{75.9$\pm$0.4} \\ 
        WL           &74.8$\pm$0.2 &72.3$\pm$3.4 &47.0$\pm$0.5 &68.8$\pm$0.4 &46.5$\pm$0.2 &80.7$\pm$3.0  &58.0$\pm$0.5 &80.0$\pm$0.5 &77.5$\pm$0.6 &72.9$\pm$0.6 \\
        DGK          &73.1$\pm$0.3 &67.0$\pm$0.6 &44.6$\pm$0.5 &78.0$\pm$0.4 &41.3$\pm$0.2 &87.4$\pm$2.7  &60.1$\pm$2.6 &80.3$\pm$0.5 &71.0$\pm$0.2 &73.3$\pm$0.8 \\
        \hline
        node2vec     &-            &-            &-            &-            &-            &72.6$\pm$10.2 &58.6$\pm$8.0 &54.9$\pm$1.6 &-            &57.5$\pm$3.6\\
        sub2vec      &-            &55.3$\pm$1.5 &36.7$\pm$0.8 &71.5$\pm$0.4 &36.7$\pm$0.4 &61.1$\pm$15.8 &60.0$\pm$6.4 &52.8$\pm$1.6 &-            &53.0$\pm$5.6\\
        graph2vec    &-            &71.1$\pm$0.5 &50.4$\pm$0.9 &75.8$\pm$1.0 &47.9$\pm$0.3 &83.2$\pm$9.3  &60.2$\pm$6.9 &73.2$\pm$1.8 &-            &73.3$\pm$2.1\\
        \hline
        InfoGraph    &70.7$\pm$1.1 &73.0$\pm$0.9 &49.7$\pm$0.5 &82.5$\pm$1.4 &53.5$\pm$1.0 &89.0$\pm$1.1  &61.7$\pm$1.4 &76.2$\pm$1.1 &72.9$\pm$1.8 &74.4$\pm$0.3\\
        GraphCL      &71.3$\pm$0.6 &70.8$\pm$0.8 &49.2$\pm$0.6 &82.6$\pm$1.0 &53.1$\pm$0.4 &88.3$\pm$1.3  &61.3$\pm$2.2 &68.5$\pm$0.6 &74.7$\pm$0.7 &72.9$\pm$1.0\\
        AD-GCL       &73.3$\pm$0.6 &72.3$\pm$0.6 &49.9$\pm$0.7 &85.5$\pm$0.8 &54.9$\pm$0.4 &89.7$\pm$1.0  &56.0$\pm$3.6 &69.7$\pm$0.5 &75.1$\pm$0.4 &73.8$\pm$0.5 \\
        SKR(ours)    &\textbf{76.3$\pm$0.6} &\textbf{74.9$\pm$1.0} &\textbf{50.9$\pm$0.4} &\textbf{91.3$\pm$0.8} &\textbf{55.6$\pm$0.7} &\textbf{90.5$\pm$0.5}  &\textbf{63.5$\pm$1.3} &\textbf{80.4$\pm$0.4} &\textbf{78.4$\pm$0.7} &71.7$\pm$0.4\\
        \hline
    \end{tabular}}
    \caption{Unsupervised learning performance for small molecules, bioinformatics and social network classification in TU
    datasets~\cite{morris2020tudataset} (Averaged accuracy(\%) $\pm$ std.(\%) over 5 runs).}
    \label{tab_classification}
\end{table*}

\subsection{Datasets and Settings}

We use graph classification benchmark datasets that are widely used in the existing graph representation learning approaches. we conduct experiments on 10 well-known benchmark datasets:
MUTAG, PTC-MR, NCI1, DD, PROTEINS, IMDB-B, IMDB-M, REDDIT-B and REDDIT-M. The detail is shown in Tab.~\ref{tab_datasets}. 

We closely follow the evaluation protocol of previous state-of-the-art graph contrastive learning approaches. For graph classification, we report the mean 10-fold cross validation accuracy after 5 runs followed by a linear SVM. The linear SVM is trained by applying cross validation on training data folds and the best mean accuracy is reported, the parameter C of SVM was selected from $\{10^{-3}, 10^{-2},..., 10^2, 10^3\}$. To make comparison fair, we adopt the basic setting of InfoGraph for graph classification and use the well known GNN architecture GIN. Specifically, we fix the number of GIN layers to 4 and training epochs to 10, and the initial learning rate is choosen from $\{1\times10^{-2}, 5\times10^{-3}, 1\times10^{-3} \}$, dimension of output representations is set to 128. The hyperparameter $\alpha$ in our augmentation strategy is selected from $\{1, 10, 50, 100, 500\}$.

We compare our SKR method with 9 unsupervised/self-supervised learning baselines for graph-level tasks, which include kernel based methods: SP~\cite{borgwardt2005shortest}, WL~\cite{shervashidze2011weisfeiler}, DGK~\cite{yanardag2015deep};
contextual based methods (skip-gram): node2vec~\cite{grover2016node2vec}, sub2vec~\cite{adhikari2018sub2vec}, graph2vec~\cite{narayanan2017graph2vec}; contrastive learning based methods: InfoGraph~\cite{sun2019infograph}, GraphCL~\cite{hafidi2020graphcl}, AD-GCL~\cite{suresh2021adversarial}

\subsection{Results and Observations}

The experimental results are summarized in Tab.~\ref{tab_classification}. Overall, from the table, we can see that our SKR model shows strong performance across almost all datasets. We make other observations as follows. 

SKR achieves considerable improvement over another competitive methods on social network datasets. The results demonstrate that SKR achieves highly competitive performance with up to 5.8\% relative improvement in accuracy on social network classification. Due to the property of social network datasets, the nodes of the data have no attribute, but the data has complicated structure. Thus results imply that SKR is a powerful method to extract topological information from graph with complicated structure.

The performance of traditional contrastive learning methods like GraphCL is inferior to the graph kernel methods that compute inner products between graphs on NCI1 dataset (12\% lower than graph kernel), which suggests that mutual information measurement may not suitable for processing NCI1 dataset comparing with inner product measurement. Because NCI1 dataset has a very sparse node feature (over 70\% parts are 0 for each node attribute), thus it implies that our SKR method by using structure knowledge are capable of mining representation from graph with sparse node feature.

\subsection{Sensitivity Analysis and Ablation Study}

\begin{figure*}[tb]
    \centering
    \includegraphics[width=6.8in]{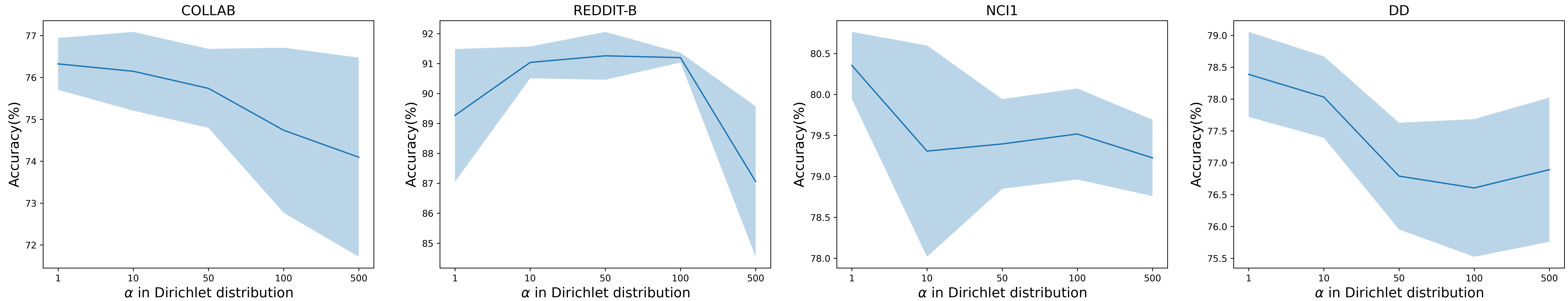}
    \vspace{-8pt}
    \caption{Sensitivity analysis on hyperparameter $\alpha$ in Dirichlet distribution on different datasets}
    \vspace{-8pt}
    \label{fig_ablation_alpha}
\end{figure*}

\begin{figure*}[tb]
    \centering
    \includegraphics[width=6.8in]{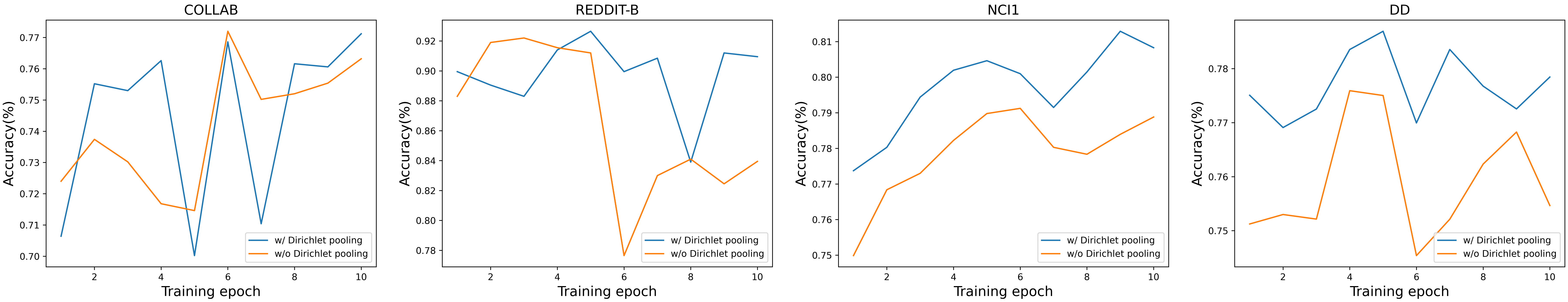}
    \vspace{-8pt}
    \caption{Ablation study on Dirichlet pooling on different datasets (with Dirichlet pooling vs without Dirichlet pooling)}
    \vspace{-8pt}
    \label{fig_ablation_augmentation}
\end{figure*}

\begin{figure}[tb]
    \centering
    \includegraphics[width=3.2in]{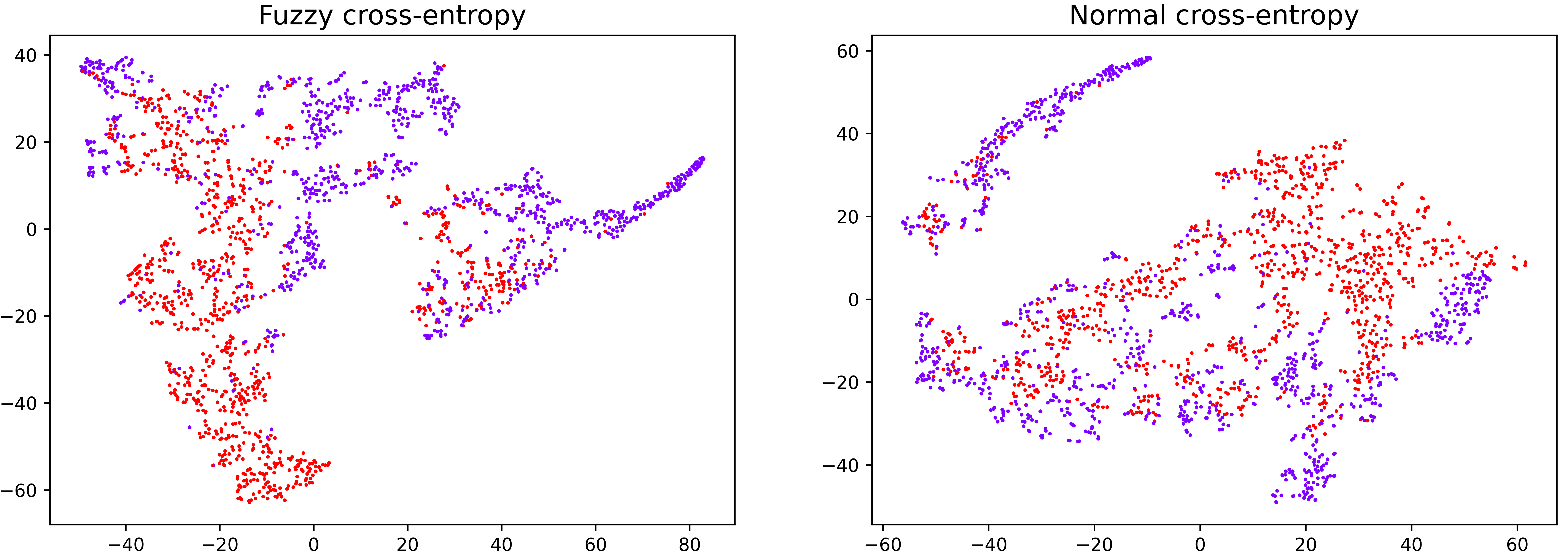}
    \vspace{-8pt}
    \caption{Ablation study on fuzzy cross-entropy on REDDIT-B dataset (Fuzzy cross-entropy vs Normal cross-entropy)}
    \vspace{-8pt}
    \label{fig_ablation_loss}
\end{figure}

We perform sensitivity analysis on critical hyperparameter $\alpha$ in our augmentation strategy as well as ablation studies on our augmentation strategy and objective function. Results show that our method is stable to perturbation of these parameters and verify the necessity of fuzzy cross-entropy and our augmentation strategy. Details are shown in following. 

\textbf{Sensitivity analysis on hyperparameter $\alpha$ in Dirichlet distribution.} In this experiment, we fix other hyperparameters and only change $\alpha$, we run 5 times for each $\alpha$ value over differnt random seed, and the accuracies is derived by settings above, then we take mean accuracy by averaging on random seeds. The results are shown in Fig.~\ref{fig_ablation_alpha}, we use 4 datasets from 3 different domains. It showns that accuracies is insensitive to the change of $\alpha$ (the fluctuations of accuracies are no more than 3\%). According to Eq.~\ref{property of Dirichlet}, we know that small $\alpha$ indicates large perturbation on semantic graph-level representation and large $\alpha$ indicates small perturbation. From Fig.~\ref{fig_ablation_alpha} we see that even under a large perturbation ($\alpha=1$) the accuracies of these 4 datasets still show strong performances.  And when the perturbation is small ($\alpha=500$), the accuracies decrease a little bit, it is consistent with our analysis of Dirichlet pool, because large $\alpha$ means tiny perturbation on semantic graph-level representation, thus the augmented representations has no difference with original semantic representations which means there is no augmentation in semantic space. 

\textbf{Ablation study on our augmentation strategy.}
In this experiment, we perform ablation study on two schemes for graph-level representation learning, with Dirichlet pooling and without Dirichlet pooling. We fix other conditions and only tuning the existence of Dirichlet pooling, then plot the figure of accuracy during training process. The results are shown in Fig.~\ref{fig_ablation_augmentation}, we use 4 datasets from 3 different domains. The results show that Dirichlet pooling is important in our SKR method, and confirm that Dirichlet pooling refine the embedding structure knowledge to get better embedding representations by using augment the semantic structure knowledge.

\textbf{Ablation study on our objective function.}
In this experiment, we perform ablation study on two schemes for graph-level representation learning, using fuzzy cross-entropy and using normal cross-entropy. We fix other conditions and only change the form of loss, then show the 2D visualization by using TSNE. The results are shown in Fig.~\ref{fig_ablation_loss}, we use REDDIT-BINARY dataset. The results show that fuzzy cross-entropy is important in our SKR method and indicate that using fuzzy cross-entropy can make the output more discriminative compared with normal cross-entropy. The results can be explain by the reason that fuzzy cross-entropy not only can attract samples with same semantic but also repel samples with different semantic, however the normal cross-entropy only attract samples with same semantic, thus the repel force is negligible for normal cross-entropy, so the output samples derived from normal cross-entropy are tightly entangled.

\section{Conclusion and Future Work}

In this paper, we propose Structure Knowledge Refinement (SKR) to learn unsupervised graph-level representations. The objective function of SKR is fuzzy cross-entropy which can automatically attract samples with same semantic and repel samples with different semantic. The augmentation strategy in SKR is Dirichlet pooling which can naturally preserve semantic. Both fuzzy cross-entropy and Dirichlet pooling play the important role in refining embedding structure Knowledge to get better embedding representations. We conduct experiments on graph classification tasks to evaluate our method. Experimental results show that SKR is competitive with state-of-the-art methods. There are many research works on semi-supervised learning on image data, but few of them focus on semi-supervised learning for graph structured data. In the future, we aim to explore semi-supervised frameworks designed specifically for graphs and apply our SKR method to node-level representation learning.

\clearpage

\bibliographystyle{named}
\bibliography{citation.bib}

\end{document}